\documentclass{article}
\usepackage{ijcai26}
\usepackage{graphicx}
\usepackage{times}
\usepackage{soul}
\usepackage{url}
\usepackage[hidelinks]{hyperref}
\usepackage[utf8]{inputenc}
\usepackage[small]{caption}
\usepackage{graphicx}
\usepackage{amsmath}
\usepackage{amsthm}
\usepackage{booktabs}
\usepackage{algorithm}
\usepackage{algorithmic}
\usepackage[switch]{lineno}
\usepackage{microtype}
\usepackage{subcaption}

\title{AlgoSimBench: \\ Identifying Algorithmically Similar Problems  for Competitive Programming}

\author{
Jierui Li
\and
Raymond Mooney\\
\affiliations
The University of Texas at Austin\\
\emails
\{jierui, mooney\}@cs.utexas.edu
}

\begin{document}

\maketitle

\begin{abstract}
Recent reasoning-enhanced Large Language Models (LLMs) have achieved promising results in solving complex competitive programming problems. However, it remains unclear whether these reasoning abilities generalize to relevant tasks, like identifying algorithmically similar problems (ASPs). We introduce AlgoSimBench, a benchmark of 402 multiple-choice questions curated in an adversarial setting: each given reference problem is paired with one algorithmically similar problem and three distractors that are semantically close but algorithmically dissimilar. This design forces models to rely on algorithmic reasoning rather than superficial textual cues. Our evaluation shows that LLMs consistently struggle under this setting. To address this gap, we propose Attempted Solution Matching (ASM), which leverages LLM-generated solution attempts to assess similarity, yielding an average accuracy improvement of 9\% across models. Beyond LLM evaluation, AlgoSimBench also probes code retrieval methods; when combined with BM25, ASM achieves an additional 11.8\% gain over state-of-the-art embedding models. AlgoSimBench offers a challenging testbed that facilitates future studies on LLMs and retrieval methods.
\end{abstract}


\begin{links}
    \link{Code}{https://github.com/lijierui/AlgoSimBench}
    \link{Datasets}{https://huggingface.co/datasets/JerryL/AlgoSimBench}
    \link{Extended version}{https://arxiv.org/abs/2507.15378}
\end{links}
\section{Introduction}

Large Language Models (LLMs) trained on both natural language and code have shown remarkable capabilities in automating various aspects of code generation and software engineering~\cite{chen2021evaluating,jimenez2024swebench}. 
Although solving competitive programming (CP) problems has become a popular and challenging benchmark to evaluate LLMs’ code reasoning abilities, it remains largely focused on code generation~\cite{li2022competition,shi2024can}. This raises an open question: can models trained on problem-solving generalize their reasoning to more abstract or alternative forms of algorithmic understanding? 

Recent works have highlighted several limitations of LLMs’ code reasoning abilities. Gu et al.\shortcite{gu2024cruxeval} find that LLMs often struggle to predict program outputs given specific inputs. Similarly, Wei et al.\shortcite{wei2025equibenchbenchmarkingcodereasoning} shows that LLMs struggle to determine whether two programs are functionally equivalent. 
In this paper, we propose to study an underexplored but crucial subskill that underlies many forms of code reasoning: the ability to identify algorithmically similar problems.

\begin{figure*}
    \centering
    \includegraphics[width=0.9\linewidth]{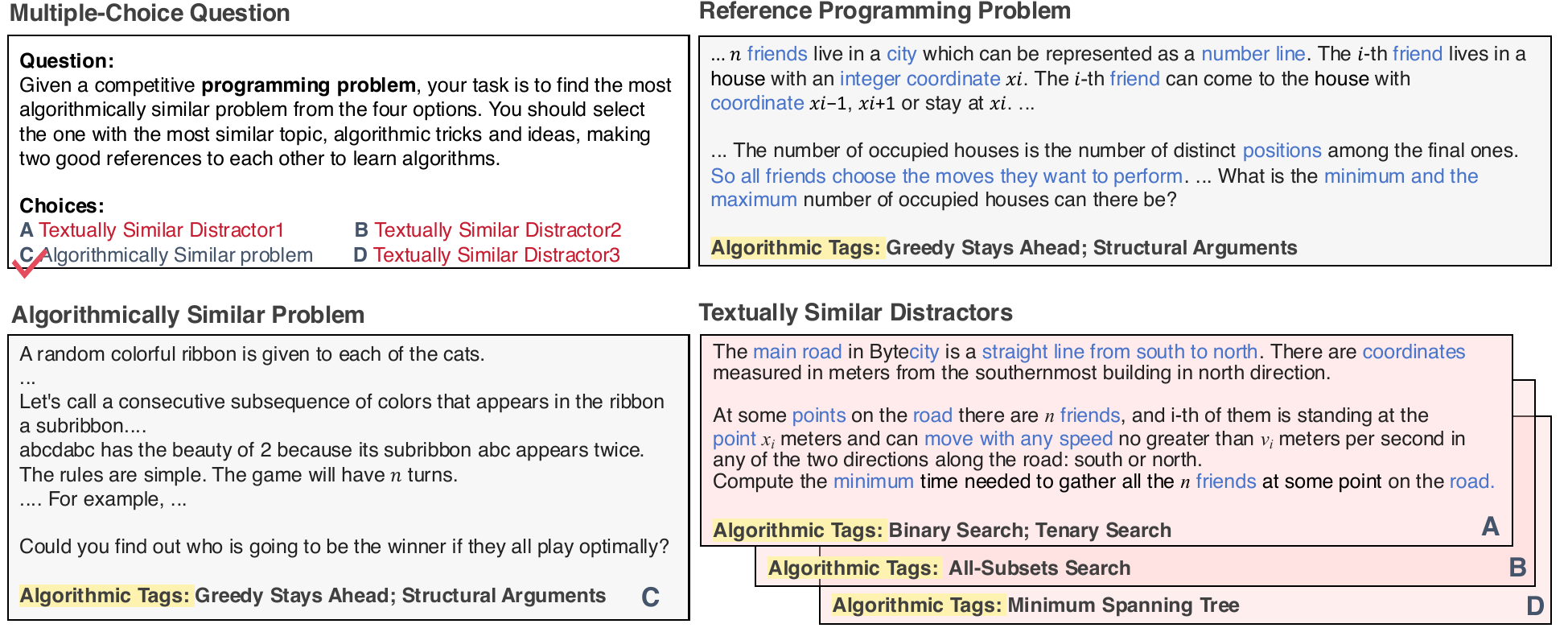}
    \caption{An example from AlgoSimBench, with one algorithmically similar problem and a textually similar distractor as an example. The blue-highlighted text makes the reference problem and the distractor look similar.}
    \label{fig:example_data}
\end{figure*}

In the context of programming and problem-solving, ``algorithmically similar problems'' (ASPs) are those that—despite differences in context, surface wording, or background story—can be solved using similar algorithmic strategies(Figure \ref{fig:example_data}). This notion of similarity goes beyond superficial textual resemblance; it lies in the shared algorithmic structure, solution path, and problem-solving ``tricks'' required. 

To curate AlgoSimBench, we collect and filter human-expert-annotated CP problems with fine-grained labels from 4 CP websites, from which we develop a set of 402 multiple-choice questions (MCQs). Each MCQ requires identifying an ASP for a reference problem from 4 options: one ASP and 3 algorithmically-dissimilar but textually similar distractors. 

We intentionally design the task to challenge superficial textual similarity analysis, emphasizing true structural algorithmic reasoning. Inspired by adversarial benchmarks such as Winograd and WinoGrande~\cite{Levesque2011TheWS,sakaguchi2019winograndeadversarialwinogradschema}, AlgoSimBench deliberately inverts the usual correlation between surface semantics and problem similarity in terms of solving. By making distractors highly semantically similar yet algorithmically different, we enforce that success does not come from shallow lexical overlap but requires deep reasoning about problem-solving and algorithmic properties. 

AlgoSimBench can be applied to evaluate LLMs in both an end-to-end selection setting and under a retrieval-based setting (see Sec. \ref{para: eval_setting}). AlgoSimBench can encourage the development of better LLMs to reason independently of superficial textual features, as well as better embedding models to represent entailed logical features in text.

Our initial study shows that program solutions are better for identifying ASPs than natural-language problem descriptions. This motivates our proposed method: attempted solution matching (ASM), in which LLMs first generate an attempted natural-language or programming-language solution for each candidate problem, and then these solutions are compared instead of problem statements to identify ASPs. ASM leverages attempted solutions as a bridge to problem similarity, yielding robust gains on AlgoSimBench for both End-to-End and Retrieval-Based selection.

Finally, we found that ASM can help improve the selection of exemplars for In-Context Learning (ICL) in the context of competitive programming. Unlike previous methods~\cite{xiong2026dqloredualquerieslow,shi-etal-2022-natural}, ASM does not require any information from human-generated oracle solutions, and yet achieves comparable performance.

\section{Related Works}
\paragraph{Solving Competitive Programming Problems with LLMs.} Competitive programming presents a significantly more challenging landscape than general code generation tasks~\cite{chen2021evaluating,austin2021program,evalplus}, as it often demands deep algorithmic reasoning, problem decomposition, and planning before implementation. Li et al.\shortcite{li2022competition} first framed competitive programming as a benchmark for evaluating language-and-code models and it was later advocated as a way to evaluate LLMs by Huang et al.~\shortcite{huang2024competitionlevelproblemseffectivellm}.
CodeLMs usually pre-train and post-train LLMs on real or synthesized code data, relying on supervised fine-tuning ~\cite{rozière2024codellamaopenfoundation,guo2024deepseekcoderlargelanguagemodel,hui2024qwen25codertechnicalreport} or policy optimization with program correctness as the reward~\cite{deepseekai2025deepseekr1incentivizingreasoningcapability}. 
Multi-agent frameworks ~\cite{islam2024mapcoder,li-etal-2025-codetree} assign separate agents to reasoning, implementation, and debugging, reflecting how humans tackle problems step by step. 
Shi et al.\shortcite{shi2024can} shows that selecting exemplar problems and solutions from top human coders for ICL provides significant gains, suggesting that good ICL examples can help LLMs better solve programming problems.

\paragraph{Code/Text Retrieval and Code Similarity.}
Early code similarity detectors relied on lexical overlap and syntax. Tools like MOSS~\cite{schleimer2003winnowing} used token matching to flag plagiarism. However, lexical and syntax match could be easily evaded by superficial code modifications~\cite{zakeri2023systematic}. To overcome these limitations, semantic code search techniques emerged~\cite{gu2018deep,feng-etal-2020-codebert}. 
Suresh et al.~\shortcite{suresh2025cornstackhighqualitycontrastivedata} construct code retrieval contrastive datasets through mining hard negatives. However, Wang et al.~\shortcite{wang2025coderagbenchretrievalaugmentcode} has found that retrievers struggle to fetch relevant and helpful context for code generation. 
Wei et al.\shortcite{wei2025equibenchbenchmarkingcodereasoning} showed that LLMs can perform poorly on the deceptively simple task of determining functional equivalence between two programs. Unlike prior works on code similarity studying snippet-level functionality (e.g., sort an array of floats), AlgoSimBench focuses on higher-level logical similarity, which lies in the problem structure and potential solutions. 

\paragraph{Disentangling Logical Reasoning in Context.}
Disentangling semantic understanding from logical or formal/symbolic reasoning has been explored across a range of tasks and domains. Zhong et al.\shortcite{zhong2023disentangling} separated semantic representation from multi-step reasoning by decomposing general QA into two dedicated modules. Hua et al.~\shortcite{hua2024disentanglinglogicrolecontext} introduced ContextHub to analyze the role of context in logical problems, showing that reasoning performance is strongly affected by natural-language–rich context.~\cite{semeval_task11} examined this issue in syllogistic reasoning by injecting content bias, benchmarking whether LLMs can judge an argument’s formal validity independently of its plausibility. Related efforts in math word problem solving similarly aim to remove linguistic confounds by transforming problems into a more purely logical form~\cite{calais2025disentangling}. To the best of our knowledge, AlgoSimBench is the first work to tackle this disentanglement in the code-and-algorithms domain, and the first to explicitly contrast textual and algorithmic similarities to enforce deeper logical reasoning.

\section{AlgoSimBench Dataset}
AlgoSimBench is a challenging benchmark, designed to evaluate LLMs' algorithmic-reasoning ability beyond code generation, code embedding  and code retrieval. 
It is collected from competitive programming problems, written, solved, labeled, and categorized by human experts. We collect problems and expert annotations from competitive programming communities, constructing 402 MCQs for ASP identification:
Each question contains a given reference problem and 4 options, one algorithmically similar problem and three algorithmically dissimilar distractors. In this section, we introduce how we collect the labeled problems and the method used to construct adversarial MCQs.

\subsection{Data Sources and Statistics}
\begin{figure}[t]
  \centering
  \begin{subfigure}[t]{0.45\textwidth}
    \centering
    \includegraphics[width=\linewidth]{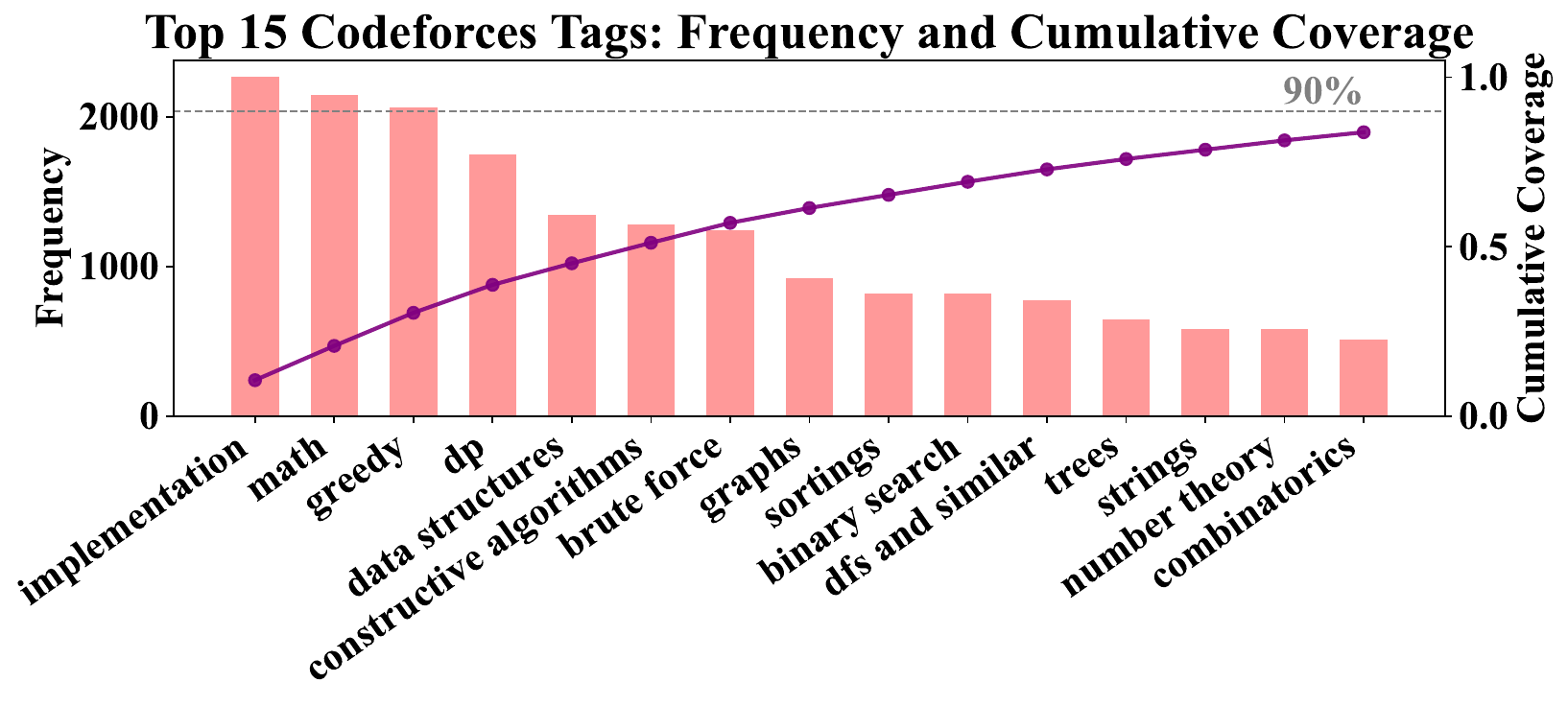}
    \caption{Top 15 Codeforces broad algorithm tags.}
    \label{fig:cf-tags}
  \end{subfigure}
  \hfill
  \begin{subfigure}[t]{0.48\textwidth}
    \centering
    \includegraphics[width=\linewidth]{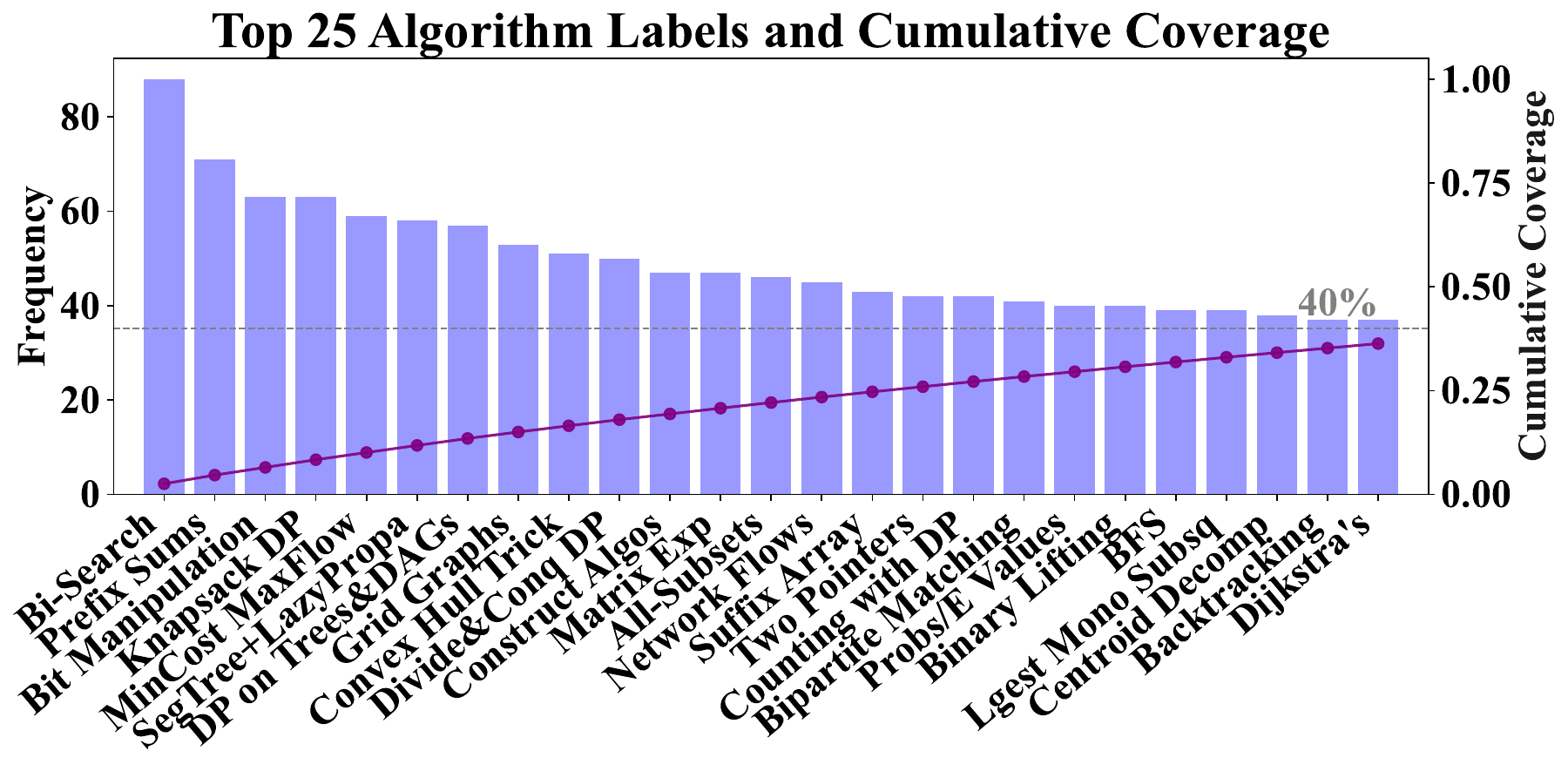} 
    \caption{Top 25 fine-grained labels in our dataset.}
    \label{fig:labeled-tags}
  \end{subfigure}
  \caption{Comparison between Codeforces tag distribution and our fine-grained tag distribution.}
  \label{fig:tag-comparison}
\end{figure}

Many competitive programming platforms provide algorithm tags, but these are usually too general (e.g., dynamic programming) to fully capture the solution strategy. To identify ASPs, we need more specific tags, such as Longest Increasing Subsequence or 0/1 Knapsack, characterizing a specific technique needed to solve the problem. A comparison between the distributions of AlgoSimBench's labels and Codeforces' tags are given in Figure \ref{fig:tag-comparison}. From four different CP community websites, we collect problems with corresponding correctness-verified human-written solutions (subsequently called \textit{oracle solutions}) and fine-grained algorithmic labels. 
We manually unified labels from these different sources and removed duplicates and overly broad or ambiguous tags, resulting in 1,317 cleanly-labeled problems. The final 402 MCQs utilized 903 problems and 231 labels from the filtered set. 

\subsection{Adversarial Task Design}
As Jain et al. and Wei et al.\shortcite{jain-etal-2021-contrastive,wei2025equibenchbenchmarkingcodereasoning} found, language and code models are sensitive to functionality-preserving attacks and struggle to disentangle program logic from natural language semantics. 

Competitive programming problems usually contain a background story, connecting the algorithms to a real-life scenario. 
We intentionally invert the correlation between algorithmic similarity and semantic textual similarity. That is, algorithmically similar problems should differ as much as possible in wording and surface descriptions, while dissimilar problems should appear similar in text but differ in algorithmic structure. Specifically:
The similar option shares the closest match in problem topics and algorithmic approach but differs significantly in text and overall linguistic meaning.
The distractors belong to different algorithmic categories but are chosen to be textually close to the original problem.
The idea underlying this setting is that \textit{models should be able to ignore features irrelevant to problem solving and base their judgment of algorithmic similarity solely on problem logic and structure.}

\subsection{Multiple-Choice Question Construction}
We obtain a problem set, $S$, each with a problem statement, $s_i$, a  corresponding solution program, $c_i$, and a set of algorithmic labels $ Y_i =\{y_1, \cdots, y_k\}$
We create multiple-choice questions as follows: Given a reference CP problem $p \in S$, we find an algorithmically similar problem, $q \in S$, and three algorithmically dissimilar problems $\{d_1, d_2, d_3\} \subset S$. 

\paragraph{Algorithmic Similarity Filter.} To ensure algorithmic similarity between the reference problem $p$ and the similar choice $q$, they should have the exact same set of labels $Y_p = Y_q$. On the other hand, the distractors $d_i$ should have no label overlap with the reference problem $p$. To further enforce the minimal algorithmic similarity of each $d_i$, we apply the following conditions to $d_i$'s labels: 1. Any $Y_{d_i}$ does not share label {\bf sub}categories with $ Y_p$; 2. Any $Y_{d_i}$' is lexically dissimilar with $Y_p$;
We then manually filter and remove edge cases. 

\paragraph{Textual Similarity Filter.}
Once we select candidates for the correct option as well as distractors, we ensure textual dissimilarity and similarity as follows. For the correct ASP option, $q$, among all candidates, we select the problem with the lowest textual similarity to $p$. For the distractors $d_i$, we select the $n$ options with the highest textual similarity to $p$, ensuring $Sim(p, d_i) > Sim(p, q)$. We compute textual similarity using a dense embedding model~\cite{lewis2020bart}.

\section{Methods}

Given the definition of algorithmic similarity, we hypothesize that it is better reflected in the solution code than in the problem statement — a finding supported by Shi et al.\shortcite{shi2024can}.
 
\subsection{Hypothesis: solutions manifest algorithmic features}
\label{sec:retri}
To test this hypothesis, we apply a direct similarity-based retrieval approach where the goal is to retrieve ASPs (from the options in our MCQ problems) using either problem statements or oracle solution code as the problem representation. We treat the 4 options $q, d_1, d_2, d_3$ as a mini corpus and use the reference problem $p$ as the query. Retrieval is successful when $q$ is the highest-ranking option. A variety of dense and sparse retrieval methods were tested using this setting. We use cosine similarity to calculate the similarity scores. As shown in Table~\ref{tab:retrieve_comparison}, retrieval based on oracle code solutions consistently achieves higher accuracy, significantly better than random. 
This supports our hypothesis: Solutions explicitly encode algorithmic features that remain implicit or latent in the problem.

\begin{table}[t]
    \centering
    \small
    \setlength{\tabcolsep}{4pt}
    \begin{tabular}{lcc}
        \toprule
        & \textbf{Statement} & \textbf{Solution}  \\
        \midrule
        BM25~\cite{robertson1995okapi}& 12.7 & 32.6  \\
        BART~\cite{lewis2020bart} & 13.2 & 33.3    \\
        CodeBert~\cite{feng-etal-2020-codebert} & 7.0 & 36.3  \\
        GraphCodeBert~\cite{guo2021graphcodebert} & 6.7 & 35.6  \\
        
        CodeSage-v2~\cite{zhang2024code} & 15.2 & 39.1 \\
        SFR-Code~\cite{liu2024codexembedgeneralistembeddingmodel} & 21.6 & 39.6  \\
        Jina-Code-V2~\cite{günther2024jinaembeddings28192token} & 10.4 & 42.8 \\
        CodeRankEmb\cite{suresh2025cornstackhighqualitycontrastivedata}  & 7.7 & 36.3  \\
        
        \bottomrule
    \end{tabular}
    \caption{AlgoSimBench MCQ Retrieval Accuracies (\%) across different retrieval methods. Cosine similarity is used for all dense models. }
    \label{tab:retrieve_comparison}
\end{table}

\begin{figure}
    \centering
    \includegraphics[width=\linewidth]{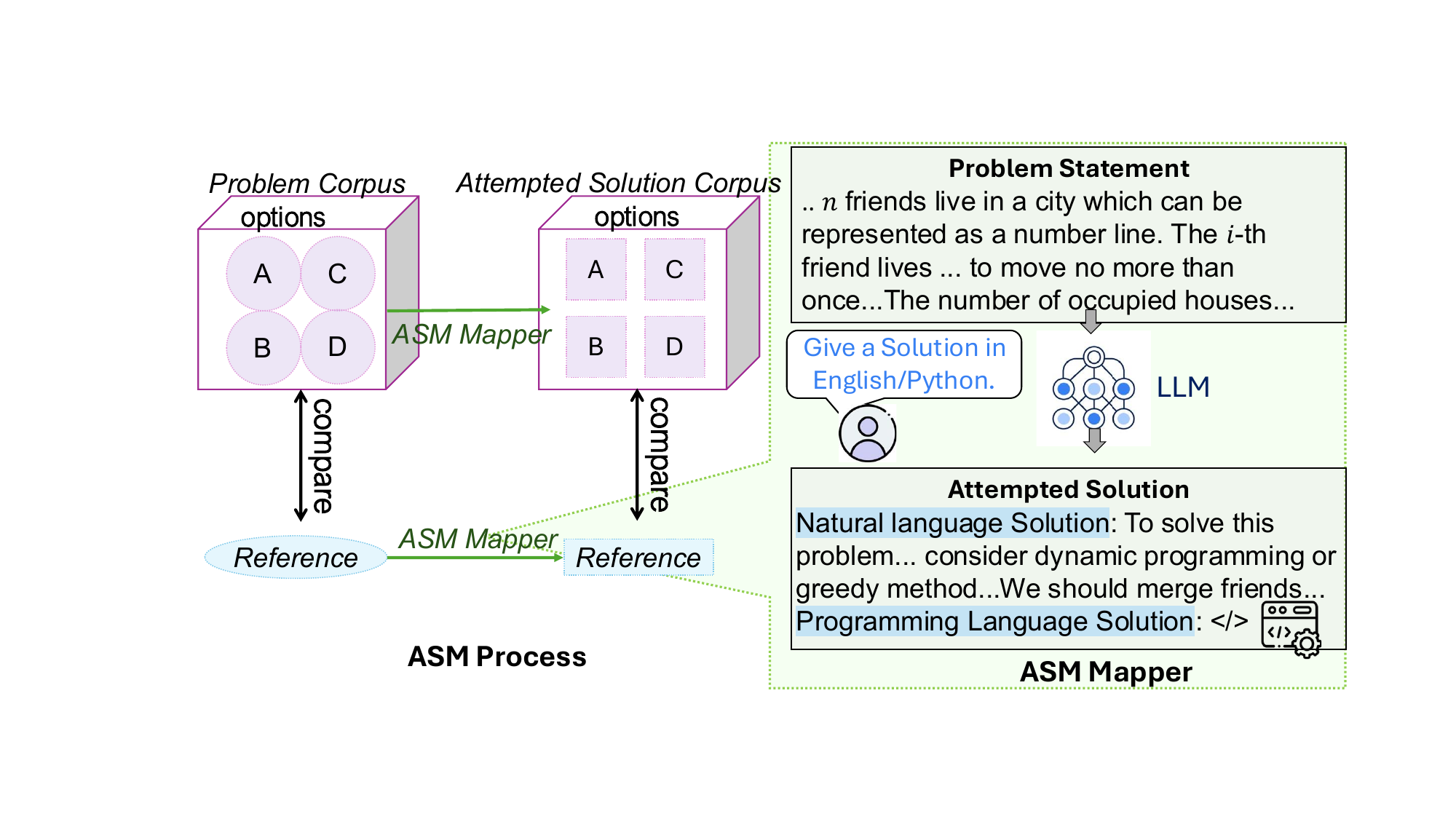}
    \caption{Illustration of Attempted Solution Matching. Both the given reference problem (reference) and options (corpus of problems) are mapped to attempted solutions by instructing an LLM to solve the problems. }
    \label{fig:method-as}
\end{figure}

Since oracle solutions are generally not available in real-world use cases,
we propose Attempted Solution Matching (ASM)—-a method that uses an LLM to generate a solution attempt for each problem, and then compares these attempted solutions to identify ASPs. Figure \ref{fig:method-as} illustrates the approach.  

\subsection{Attempted Solution Matching}

Episodic Retrieval~\cite{shi2024can} uses similarity among solution programs to identify similar problems, but it compares LLM-generated code for the query problem with oracle-solution code for the potential retrieval targets. 
This assumes that the initial solution contains the correct algorithmic signal. However, this breaks down when the LLM selects the wrong approach where such errors can accumulate, decreasing LLMs' capacity to identify algorithmically similar problems. 

To mitigate this mismatch between initial attempts and oracle solutions, we draw insights from the ``generate-to-retrieve'' paradigm~\cite{gao-etal-2023-precise} and propose matching first-attempt solutions to other first-attempt solutions. 
Our method comes with a simple assumption: when attempting to solve a problem, the algorithmic ideas and problem-solving strategies are naturally reflected in both the model’s natural language reasoning and its generated code, revealing the underlying methods being applied. As a result, algorithmically similar problems tend to produce similar solution attempts.

In ASM, each problem statement (i.e., problem description) is first mapped to a first-attempt solution using an LLM, and ASPs are identified by evaluating the similarities between attempted solutions. 
The attempted solution can take two forms: A natural language explanation of the solution strategy (NL-solution), denoted as ASM-NL; or a solution written in a programming language (PL-solution), ASM-PL. 
While both should contain algorithmic information, ASM-NL can include content on problem analysis, strategy exploration, algorithm selection, and problem solving; on the other hand, ASM-PL focuses on the complete implementation, whose content is rich in details but lacking in problem analysis. 

\section{Experiments}
In this section, we start with evaluating LLMs' abilities to identify ASPs with AlgoSimBench under End-to-End Selection. Experiments are conducted to analyze how different factors affect LLMs' performance. Next, we explore using AlgoSimBench to test the abilities of various retrieval methods under the Retrieval-Based Selection setting. We evaluate several baselines as well as ASM-NL and ASM-PL under both settings. Finally, we apply ASM to in-context learning to retrieve better examples for aiding the generation of solutions to programming problems.

\subsection{Experimental Settings}\label{settings}

\paragraph{Datasets \& Metrics.}AlgoSimBench, consists of 402 MCQs, where we measure the accuracy as the rate of selecting the correct ASP out of 4 options. To evaluate the performance of ICL exemplar selection, we test ASM and various baselines on USACO~\cite{shi2024can}, a dataset with 307 "olympiad" competitive programming problems. We calculate the increase in the pass@1 solution rate~\cite{chen2021evaluating} over random exemplar selection for different exemplar retrieval methods. 

\begin{table*}[t]
\centering
\small
\begin{tabular}{lccll|c}
\toprule
\textbf{Model} & \textbf{Statement} & \textbf{Summary} & \textbf{ASM-NL} & \textbf{ASM-PL} & \textbf{Solution$^*$} \\
\midrule
Qwen3-0.6B-thinking & 25.8 & 26.1 & \textbf{33.3} ({$\uparrow$ 7.5}) & 31.3 ({$\uparrow$ 5.5}) & 30.6\\
Qwen3-4B-thinking & 43.3 & 41.2 & 53.0 ({$\uparrow$ 9.7})& \textbf{53.9} ({$\uparrow$ 10.6})& 59.2\\
Qwen3-8B-thinking & 47.5 & 46.3 & \textbf{58.2} ({$\uparrow$ 10.7}) & 57.0 ({$\uparrow$ 9.5})& 62.4\\
Llama3.1-8B-Instruct & 23.9 & 24.1 & \textbf{26.4} ({$\uparrow$ 2.5}) & 24.4 ({$\uparrow$ 0.3}) & 29.6 \\
GPT-4o-mini       & 35.6 & 35.8 & \textbf{43.8} ({$\uparrow$ 8.3}) & 42.5 ({$\uparrow$ 7.0}) & 54.4 \\
GPT-4o            & 41.5 & 38.1 & \textbf{53.2} ({$\uparrow$ 11.7}) & 53.0 ({$\uparrow$ 11.5}) & 63.4 \\
o3-mini-medium    & 65.9 & 63.4 & 74.4 ({$\uparrow$ 8.5})       & \textbf{75.1} ({$\uparrow$ 9.2}) & 72.6 \\
Deepseek-R1       & 63.7 & 57.7 & 69.2 ({$\uparrow$ 5.5})       & \textbf{70.4} ({$\uparrow$ 6.7}) & 70.6 \\
Deepseek-V3       & 55.2 & 53.2 & \textbf{64.9} ({$\uparrow$ 9.7}) & 62.7 ({$\uparrow$ 7.5}) & 66.7 \\
Claude-3.5-Sonnet & 44.2 & 45.0 & \textbf{54.7} ({$\uparrow$ 10.5}) & 53.0 ({$\uparrow$ 8.8}) & 70.5 \\
Gemini-2.0-Flash  & 51.2 & 48.5 & \textbf{57.9} ({$\uparrow$ 6.7})  & 55.5 ({$\uparrow$ 4.3}) & 69.7 \\
Qwen3-Coder-480B & 50.7 & 51.0 & \textbf{68.4} ($\uparrow$17.7) & 62.4 ($\uparrow$11.7) & 66.9\\
\midrule
\textbf{Avg}    & 45.1 & 44.2 & \textbf{54.8} ({$\uparrow$ 9.7})  & 53.4 ({$\uparrow$ 8.3}) & 59.7 \\
\bottomrule
\end{tabular}
\caption{End-to-End Selection Performances(Accuracy\%) on AlgoSimBench-MCQ over different models and methods. Results in bold indicate the best performing non-oracle method for each model. Absolute performance gain over the \textbf{Statement} baseline is marked with $\uparrow$. \textit{Solution} is an oracle-input setting where oracle solutions were directly provided to the model. ASMs are statistically significantly better than Statement/Summary with $p<0.01$.}
\label{tab:model_performance}
\end{table*}
\paragraph{Models.} 
For End-to-End Selection, we evaluate 12 open and closed LLMs: Qwen3-0.6B, 4B, 8B and Qwen3-Coder-480B \footnote{\url{https://huggingface.co/collections/Qwen/qwen3}}. Llama-3.1-8B-instruct\footnote{\url{https://huggingface.co/meta-llama/Llama-3.1-8B-Instruct}}, GPT-4o-mini, GPT-4o\footnote{\url{https://openai.com/index/hello-gpt-4o/}}, o3-mini-medium\footnote{\url{https://openai.com/index/openai-o3-mini/}}, Deepseek-R1~\cite{deepseekai2025deepseekr1incentivizingreasoningcapability}, Deepseek-V3~\cite{deepseekai2025deepseekv3technicalreport}, Claude-3.5-Sonnet\footnote{\url{https://www.anthropic.com/news/claude-3-5-sonnet}} and Gemini 2.0 Flash.\footnote{\url{https://deepmind.google/technologies/gemini/flash/}}
For Retrieval-Based Selection, we tested various metrics for measuring the similarity between the query and candidate answers. We used a sparse retriever, BM25~\cite{robertson1995okapi}, and cosine similarity of dense embeddings from a text embedding model BART~\cite{lewis2020bart}, code embedding models GraphCodeBert~\cite{guo2021graphcodebert},         CodeSage-v2~\cite{zhang2024code}, SFR-Code-400M\_R ~\cite{liu2024codexembedgeneralistembeddingmodel}, Jina-Code-V2 ~\cite{günther2024jinaembeddings28192token}, CodeRank Embedding~\cite{suresh2025cornstackhighqualitycontrastivedata}.

\paragraph{Baselines.}
We compare ASM to using the following baseline problem representations. 
\textbf{Statement}: Original natural-language descriptions are used to represent problems, \textbf{Summary}: To mitigate aspects unrelated to problem-solving, an LLM is first prompted to summarize and abstract the problem, minimizing narrative elements and formatting details irrelevant to the structure of the problem.  \textbf{Solution:} Each problem is replaced with a correct solution written by an expert programmer, which serves as an "oracle" upper baseline.
\paragraph{ASM Settings.}
As discussed before, for Attempted Solution Matching, we have a \textit{Natural Language} setting, \textbf{ASM-NL}, where LLMs are asked to describe the solution in English given the problem, which is later used as reference and options; a \textit{Programming Language} setting \textbf{ASM-PL}, where LLMs are asked to first generate Python-implemented solutions for the reference and options.

\paragraph{Evaluation Settings.}
\label{para: eval_setting}
We evaluate models under two task settings: 1) End-to-End Selection: where a full MCQ is presented to the LLM in a single prompt with a reference problem and four candidate options, and models are directly asked to select the most algorithmically similar problem. 2) Retrieval-Based Selection: where each MCQ is framed as a retrieval task, where the query is the reference problem and the 4 candidates form a corpus of potential retrievals. For both settings, it is considered correct if the retrieved option is the algorithmically similar one.

\subsection{End-to-End Selection}
We apply the same prompts to all End-to-End Selection experiments, with a definition of ASPs and encouragement to use chain-of-thought reasoning. The only difference is to provide problems or attempted solutions as the options.
\label{sec:end2end}
\subsubsection{Main Results}

MCQ accuracies of all seven models using End-to-End selection are given in Table \ref{tab:model_performance}. We found that none of the models perform particularly well when only given the original problem Statements. The weaker model \textit{Llama3.1-8B-Instruct} has near-random performance across all methods. Reasoning LLMs like Deepseek-R1 and o3-mini perform better than other models, achieving accuracies around 60\%. Given oracle solutions (\textbf{Solution}) instead of problem statements, LLMs can identify algorithmic similarity much better, yielding improvements as high as 26.3\%.

 Summarizing problems actually hurts the performance of all models except 2. We hypothesize that SoTA LLMs can already ignore superficial similarities when comparing programming problems. Our methods, \textbf{ASM-NL} and \textbf{ASM-PL}, consistently improve absolute performances across all models with an average improvement of 9.7\% and 8.3\%, respectively. \textbf{ASM-NL} outperforms \textbf{ASM-PL} on 10 out of 12 models, indicating that most models can generate and compare solutions described in natural language better than actual code solutions. For o3-mini-medium, both ASMs even outperform the oracle-solution baseline and give the best performances overall. The performance gaps between \textbf{Statement} and \textbf{ASM} also suggest that LLMs cannot generalize their code reasoning ability to similar tasks when they are out of a code generation context.

\subsubsection{Effects of Problem Attributes} 
We further analyzed failure cases in AlgoSimBench to answer two research questions: 1.  Do models make more mistakes on more difficult problems? 2. Are there some algorithms that are especially difficult to identify from problem descriptions?

\paragraph{Difficulty-level.} 
The Pearson correlation between a problem's Codeforces' difficulty rating and the probability that an LLM will get it wrong is \textbf{-0.15} with p-value = 0.45, indicating that models do not perform worse on ASP if the problems are harder. This highlights an important distinction: \textit{identifying a problem’s category and the appropriate algorithm is not the same as being able to solve it}. For instance, recognizing that a Segment Tree is needed does not necessarily imply the ability to implement a correct solution for range updates and queries. Some problems are easy to \textit{classify} but difficult to \textit{solve}—reinforcing the need to evaluate models beyond final solution correctness.

\begin{figure*}[tbp]
  \centering
  \begin{subfigure}[t]{0.47\textwidth}
    \centering
    \includegraphics[width=\textwidth]{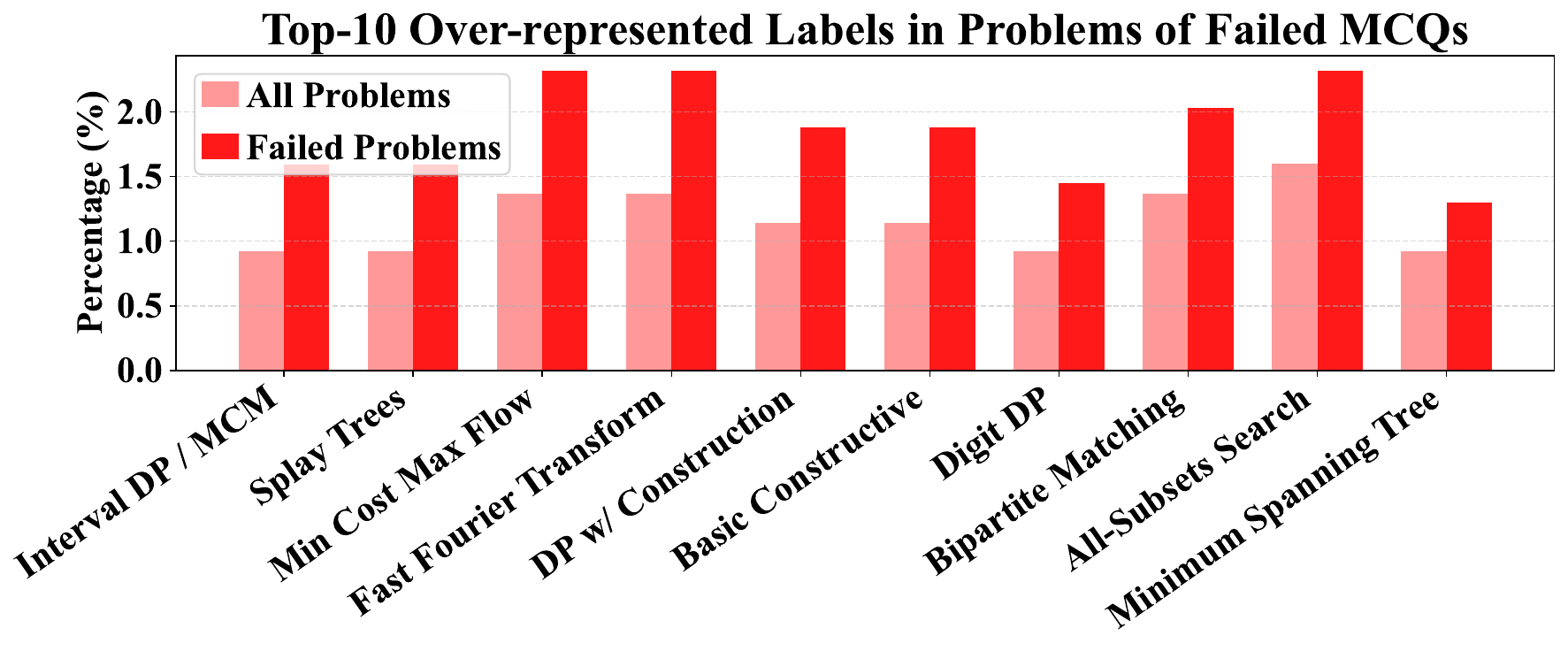}
    \caption{Top-10 over-represented (greater portion, percentage) labels in failed MCQs compared to all MCQs.}
  \end{subfigure}
  \hfill
  \begin{subfigure}[t]{0.46\textwidth}
    \centering
    \includegraphics[width=\textwidth]{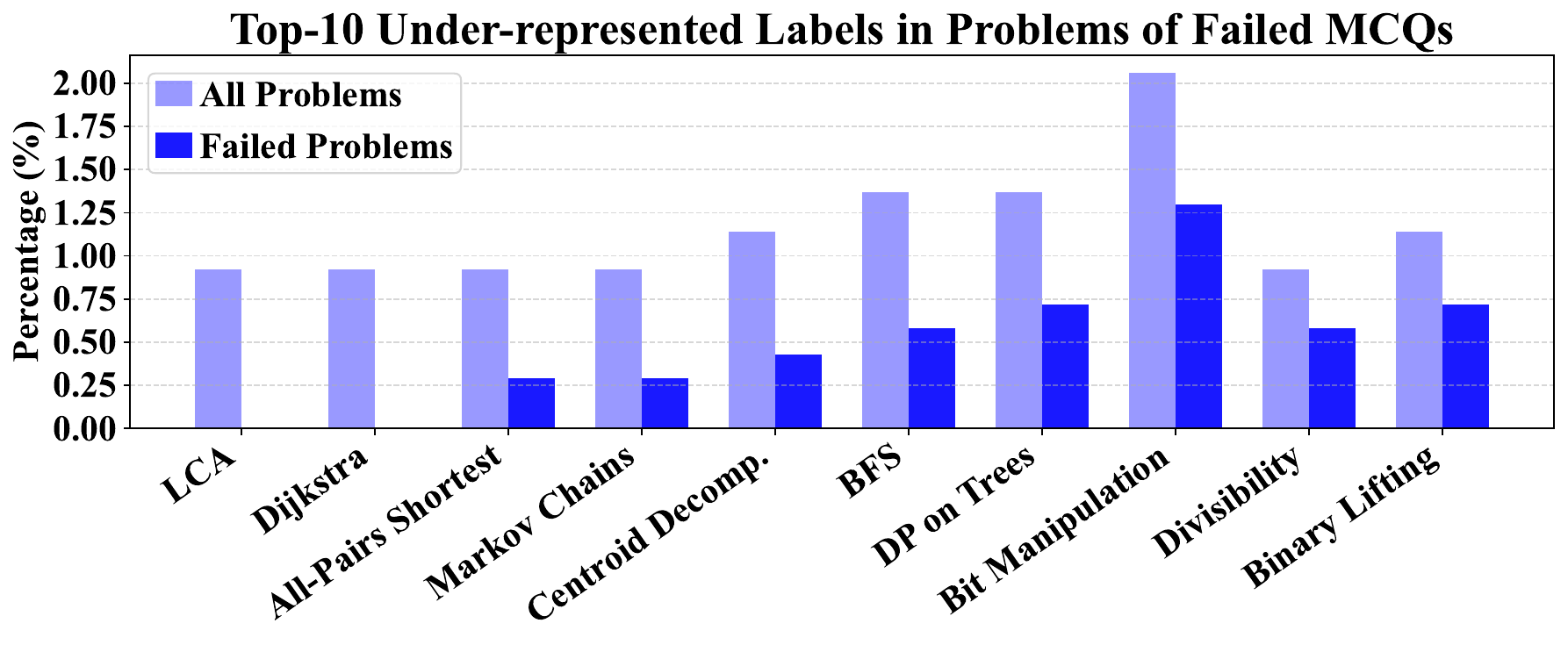}
    \caption{Top-10 under-represented (lower portion, percentage) labels in failed MCQs compared to all MCQs.}
  \end{subfigure}
  \caption{Comparison of tag distributions between problems in failed and all MCQs. Over/Under-represented is calculated by comparing the label distributions of all problems to those of failed problems.}
  \label{fig:tag_difficulty_bias}
\end{figure*}

\paragraph{Algorithm Labels.}
The over-represented and under-represented algorithmic labels characterizing problems that models tend to get wrong (i.e., highest increment and decrement in label proportion) are shown in Figure \ref{fig:tag_difficulty_bias}.
We found that some algorithmic types, though often seen as difficult (e.g., Centroid Decomposition) are not hard to identify. Some other simple problems, like shortest-path/Dijkstra, are also easy to identify. On the other hand, algorithms like DP and Fast-Fourier-Transform can be applied in very different circumstances, making such problems difficult to identify.

In a nutshell, LLMs' ability to completely solve programming problems and their ability to identify relevant algorithmic strategies are two fairly independent axes. The latter ability is more determined by a problem's type and topic than its difficulty level.

\subsubsection{Effectiveness and Efficiency of ASM}
\paragraph{Effectiveness.} As discussed earlier, models' abilities to identify ASPs are not directly reflected by their ability to fully solve the problems. With \textit{Attempted} Solution Matching, models are able to identify ASPs based on solutions that are not fully correct. To illustrate this robustness, we tested the quality of the generated programs for ASM-PL and surprisingly found that even when programs are incorrect, they can still substantially help identify algorithmically similar problems. The quality of generated programs in ASM-PL is evaluated by pass@1~\cite{chen2021evaluating}. In Table \ref{tab:pass_rate}, we find that for weaker models, with pass@1 ranging from 1.8\% to 19.6\%, matching mostly-incorrect attempts still gives significant performance gains compared to directly matching problems statements, For Qwen models, ASM performs close to matching oracle solutions. While Claude-3.5-Sonnet has similar performance to Deepseek-R1 when solving problems and matching oracle solutions, it performs much worse when matching its own generated programs. This suggests that it does not identify the correct algorithm during failed solution attempts.

\begin{table}[t]
    \centering
    \small
    \begin{tabular}{lccc}
        \toprule
        \textbf{Model} & \textbf{PL pass@1} & \textbf{ASM-PL} & \textbf{Solution} \\
        \midrule
        Qwen3-0.6B        & 1.8  & 31.3 ({$\uparrow$ 5.5}) & 30.6 \\
        Qwen3-4B          & 16.0 & 53.9 ({$\uparrow$ 10.6}) & 59.2 \\
        Qwen3-8B          & 17.9 & 57.0 ({$\uparrow$ 9.5}) & 62.4 \\
        GPT-4o-mini       & 19.6 & 42.5 ({$\uparrow$ 7.0}) & 54.4 \\
        Claude-3.5-Sonnet & 41.0 & 53.0 ({$\uparrow$ 8.8}) & 70.5 \\
        Deepseek-R1       & 52.1 & 70.4 ({$\uparrow$ 6.7}) & 70.6 \\
       
        \bottomrule
    \end{tabular}
    \caption{Comparisons of models' performances on solving problems (PL pass@1) and utilizing generated or oracle programs to identify algorithmically similar problems (ASM-PL and Solution). Numbers given are \% accuracy.}
    \label{tab:pass_rate}
\end{table}

\paragraph{Efficiency.}To illustrate the efficiency of ASM, we compare the cost (in generated tokens) of ASM-NL and ASM-PL against simply matching problem statements. We find that ASM significantly reduces  the number of reasoning tokens used to solve problems. ASM-NL costs 80\% as many tokens on average across all models. ASM-PL costs 190\% as many tokens as it needs a longer CoT to generate a complete solution than a high-level NL strategy. Even so, to \textit{obtain similar accuracy}, ASM-PL is more efficient than either relying on the built-in thinking mode of LLMs or using a larger model within the same model family.
In Table \ref{tab: asm_efficiency}, we define cost as \# gen\_tokens $\times$ \textit{unit\_price}. ASM-NL can achieve better accuracy while costing only 15\% to 26\% as much. ASM-PL also costs less than comparing statements. In other words, both ASMs can achieve better accuracy in non-thinking mode or with smaller models, yielding a lower overall cost.
\begin{table}[t]
    \centering
    \small
    \setlength{\tabcolsep}{4pt}
    \begin{tabular}{lcc}
        \toprule
        & \textbf{Accuracy\%}($\uparrow$) & \textbf{Cost}($\downarrow$)  \\
        \midrule
        Statement w/GPT-4o& 41.5 & 6.67X\\
        ASM-PL w/GPT-4o-mini & 42.5 & 1.46X \\
        ASM-NL w/GPT-4o-mini  & 43.8 & 1.00X \\
        \midrule
        Statement w/Qwen3-4B-think & 43.3& 6.23X\\
        ASM-PL w/Qwen3-4B & 46.3 & 5.49X \\
        ASM-NL w/Qwen3-4B & 47.3 & 1.00X \\
        \midrule
        Statement w/Qwen3-8B-think & 47.5 & 3.80X\\
        ASM-PL w/Qwen3-8B & 51.7 & 2.16X \\
        ASM-NL w/Qwen3-8B & 51.4 & 1.00X \\
        \bottomrule
    \end{tabular}
    \caption{Cross-model and Cross-mode performance/efficiency comparison for GPT-4o and Qwen3 models. \textit{-think} refers to model with the thinking-mode enabled.}
    \label{tab: asm_efficiency}
\end{table}

\subsection{Retrieval-Based Selection}
\label{sec:retrieval}
\begin{table*}[t]
\centering
\hspace*{-0.3cm}
\small
\scalebox{0.78}{
\setlength{\tabcolsep}{3.7pt}
\begin{tabular}{l|ccccccc|ccccccc|ccccccc}
\toprule
& \multicolumn{7}{c|}{\textbf{Summary}} & \multicolumn{7}{c|}{\textbf{ASM-NL}} & \multicolumn{7}{c}{\textbf{ASM-PL}}\\
& BM25& Bart & GCB & CR & CS & Jina & SFR & BM25& Bart & GCB & CR & CS & Jina & SFR & BM25& Bart & GCB & CR & CS & Jina & SFR \\
\midrule
GPT-4o-mini     & 25.6 & 23.4 & 20.6    & 20.1 & 19.9 & 20.6 & 28.4 &\textbf{35.3} & 29.1 & 22.4 & 23.6 & 22.9 & 24.9 & 26.1 & 34.8 & 26.1 & 32.8 & 30.3 & 24.4 & 29.4 & 30.1 \\

GPT-4o   & 25.8 & 24.6 & 26.1 & 24.9 & 19.2 & 21.6 & 28.9 & \textbf{42.5} & 30.1 & 32.1 & 28.4 & 28.1 & 31.1 & 34.6 & 35.6 & 28.8 & 29.1& 29.4 & 21.9 & 26.6 & 26.6 \\
o3-mini-medium & 39.3 & 32.6 & 31.3 & 29.1 & 29.9 & 30.6 & 34.1 & \textbf{49.0} & 35.6 & 38.3 & 40.8 & 40.3 & 40.8 & 39.3  & 48.8 & 28.4 & 39.3 & 37.3 & 35.6 & 46.3 & 35.1\\
Deepseek-V3        & 29.9 & 26.1 & 22.4 & 24.4 & 24.1 & 32.1 & 33.1 & \textbf{46.0} & 33.1 & 32.1 & 33.8 & 37.6 & 36.8 & 36.3 & 45.5 & 32.3 & 35.1 & 38.6 & 31.1 & 40.8 & 37.8 \\
Deepseek-R1 & 30.1 & 24.1 & 25.9 & 30.1 & 28.4 & 30.8 & 35.1 & \textbf{52.2} & 32.8 & 31.1 & 45.8 & 41.3 & 43.3 & 40.0 & 45.0 & 32.8 & 35.9 & 44.0 & 38.8 & 49.8 & 40.0 \\
Gemini-2.0-Flash   & 27.1 & 24.4 & 18.6 & 24.4 & 21.9 & 23.1 & 28.4 & \textbf{41.0} & 34.3 & 26.6 & 33.6 & 34.1 & 33.1 & 36.8 & 38.0 & 36.8 & 35.6 & 33.8 & 29.9 & 36.1 & 33.3 \\
Claude-Sonnet-3.5  & 29.4 & 25.1 & 25.4 & 26.1 & 25.4 & 29.1 & 32.3 & \textbf{39.6} & 28.1 & 28.6 & 32.6 & 31.6 & 36.3 & 34.6 & 35.0 & 29.1 & 29.9 & 29.4 & 26.4 & 30.8 & 33.6 \\\hline
Avg      & 29.6 & 25.8& 24.3  & 24.0 & 22.1 & 25.3 & 30.2 & \textbf{43.7} & 31.9 &30.2 & 30.4 & 30.8 & 32.4 & 33.7 &40.4 &30.6& 34.0 & 32.3 & 26.7 & 32.7 & 32.3 \\
\bottomrule
\end{tabular}}
\caption{MCQ accuracy(\%) for different LLMs used to generate summaries and attempted solutions, and different retrieval metrics: GCB=GraphCodeBert, CR = CodeRankEmbed, CS = CodeSage-v2, Jina = jina-v2-code, SFR = SFR-Embedding-Code$_{400M}$. For comparison, retrieval performance using the problem statement or oracle solution are given in Table \ref{tab:retrieve_comparison}.}
\label{tab:retrieve-based-mcq}
\end{table*}
In this setting, we treat AlgoSimBench as a retrieval problem as discussed in section \ref{para: eval_setting}. Results for retrieval based on problem statements were shown in Table \ref{tab:retrieve_comparison}. Here, we use an LLM to first map each statement to a \textbf{Summary}, an NL solution (\textbf{ASM-NL}),  and a program solution (\textbf{ASM-PL}). We evaluate how well retrieval using these different representations is able to identify the correct ASP. Results are shown in Table \ref{tab:retrieve-based-mcq}. 

\textbf{ASM-NL} again outperforms other methods. Unlike retrieval based on statements (see Table \ref{tab:retrieve_comparison}), using a Summary is better than random guessing. Since AlgoSimBench is constructed so that the textual problem-statement similarity is a misleading indicator of ASPs, summarization is helpful but still provides limited explicit information about algorithms.

An interesting finding is that simple term-matching (BM25) performs better than dense embedding models. A possible reason could be that algorithmic similarity is often better indicated by keywords like algorithmic terms or descriptions of the core idea, rather than a full implementation.

Although Retrieval-Based Selection performs significantly worse than End-to-End Selection, it scales much better to retrieving ASPs from a large corpus.
End-to-end selection requires fitting all potential retrievals into an LLM input window and is difficult to scale beyond choosing from only four options in an MCQ. In contrast, retrieval-based selection only requires running an LLM on individual problems once to produce more effective problem representations that can then be efficiently retrieved from a large corpus using standard IR techniques (such as BM25) or dense retrieval with compressed text representations~\cite{izacard2022unsuperviseddenseinformationretrieval,9942356}. 

\subsection{ICL Exemplar Selection with ASM}

Since ASM can help find algorithmically similar problems, we apply it to enhance code generation. Shi et al.~\shortcite{shi2024can} showed that including in-context exemplars of solved problems can enhance LLMs' abilities to solve competitive programming problems. They proposed Episodic Retrieval, using an LLM's solution along with the problem statement to retrieve the most similar human solutions using BM25. Then, the retrieved problem is included as an ICL exemplar, hopefully improving the ability to solve the given problem. ASM, on the other hand, directly matches LLM attempted solutions for both the given problem and problems in the corpus. Utilizing their methodology, we explored how ASM could be used for improved selection of ICL exemplars.

\begin{table}[t]
\centering 
\small
\begin{tabular}{ccc}
\toprule
Exemplar Selection            & GPT-4o-mini & GPT-4o \\
\midrule
w/o Exemplar         &   9.4\%     & 17.3\%  \\
Retrieve w/Random          & 10.4\%    & 17.9\% \\
Retrieve w/Statement  &  11.4\%&  16.9\%\\
Episodic Retrieval$^*$  & 12.4\%  &  18.6\%     \\
Retrieve w/ASM-NL & \textbf{13.7\%}  &    19.2\%    \\
Retrieve w/ASM-PL & 13.0\% &    \textbf{19.9\%}  \\

\bottomrule
\end{tabular}
\caption{ICL enhanced by different exemplar selection methods. Results are 1-shot Pass@1 on the USACO Benchmark. *Episodic Retrieval requires human-generated oracle solutions.}
\label{tab:icl}
\end{table}

We also applied two baselines that select either a \textit{Random} exemplar or retrieve the most similar exemplar using problem statements. Table \ref{tab:icl} shows that with the same Retriever (BM25, which outperformed various dense-embedding methods), ASM methods achieve the best pass@1~\cite{chen2021evaluating} performance. While the absolute improvement is modest, this might be limited by how much performance gain ICL with a single example can bring to competitive programming~\cite{Tang_2023,patel2024evaluatingincontextlearninglibraries}.


\section{Conclusion and Discussion}
This paper introduced AlgoSimBench, a novel benchmark designed to evaluate models’ ability to reason about algorithmic similarity between competitive programming problems. This benchmark: 1) provides a focused evaluation of the algorithmic reasoning abilities of LLMs, decoupled from full solution generation; and 2) enables the study of retrieval methods that go beyond surface-level textual similarity, instead capturing deeper structural and problem-solving-related semantics.

We propose Attempted Solution Matching (ASM), which leverages LLM-generated first-attempt solutions—either in natural language or code—to better reflect the core algorithmic aspects of each problem. In both end-to-end selection and retrieval settings, ASM improves the accuracy of identifying ASPs and improves exemplar selection for in-context learning. 

Some key findings were: 1) the performance gap between comparing statements and attempted solutions suggests limited generality of the algorithmic reasoning skills learned from code generation alone;
2) Identifying ASPs depends more on algorithm type rather than problem difficulty, making it a separate axis from problem-solving;
3) BM25 outperforms dense retrievers, indicating that keyword signals capture algorithmic features more effectively than generic semantic embeddings.

We hope AlgoSimBench inspires future work on identifying algorithmic similarity, code retrieval, and retrieval-augmented reasoning.

\bibliographystyle{named}
\bibliography{ijcai26}
\appendix
\section{Appendix}

\subsection{Dataset Sources \& Statistics}
\label{apd:data}
Our data was collected from four websites for competitive programming participants and learners, namely:
\begin{itemize}
    \item The Ultimate Topic List: \url{https://youkn0wwho.academy/topic-list}
    \item CP Algorithms: \url{https://cp-algorithms.com/}
    \item ProgVar: \url{https://progvar.fun/}
    \item USACO-Guide \url{https://usaco.guide/}
\end{itemize}
1,522 problem descriptions and solutions were collected from the following competitive programming websites:
Codeforces (\url{https://codeforces.com/}), AtCoder(\url{https://atcoder.jp/}) and CodeChef(\url{https://www.codechef.com/}).

We filtered out problems with broad, ambiguous, or duplicate annotations, giving a final set of 1,317 problems. Detailed information on these problems is available in Table \ref{tab:problem_stat}. Not all these problems were selected for the final MCQs, some were ignored due to the lack of similar problems or distractors or failing textual similarity thresholds.  However, the full set serves as a good source of programming problems for other tasks such as algorithmic tagging. 

\begin{table}[h]
\centering 
\small
\setlength{\tabcolsep}{3.7pt}
\begin{tabular}{ccccc}
\hline
 & Codeforces & AtCoder & CodeChef & total\\
\# problems($s_i$) &1114 &182 &18 &1317 \\
\# Tokens/$s_i$ & 507 & 346 & 468& 489\\
\# Solutions/$s_i$ & 1 & 1& 1& 1\\
\# Tags/$s_i$ &1.12 & 1.08& 1.0&1.11 \\
\hline
\end{tabular}
\caption{Statistics of Programming Problems in AlgoSimBench. 
}
\label{tab:problem_stat}
\end{table}

Finally, 903 problems were selected for use in the MCQs, we excluded a portion of problems with high-frequency tags like binary search, to enhance the diversity of the chosen problems. Some statistics on these problems are given in Table \ref{tab:mcq-problem_stat}.

\begin{table}[h]
\centering 
\small
\begin{tabular}{cc}
\hline
 & AlgoSimBench MCQ\\
\# questions & 402 \\
\# fine-grained tags & 231 \\
\# categorized tags & 204\\
Avg \# Tokens/question & 2681\\
\hline
\end{tabular}
\caption{Statistics of MCQs in AlgoSimBench. 
}
\label{tab:mcq-problem_stat}
\end{table}

\subsection{Algorithm Tags}

\begin{figure*}
    \centering
    \includegraphics[width=0.8\linewidth]{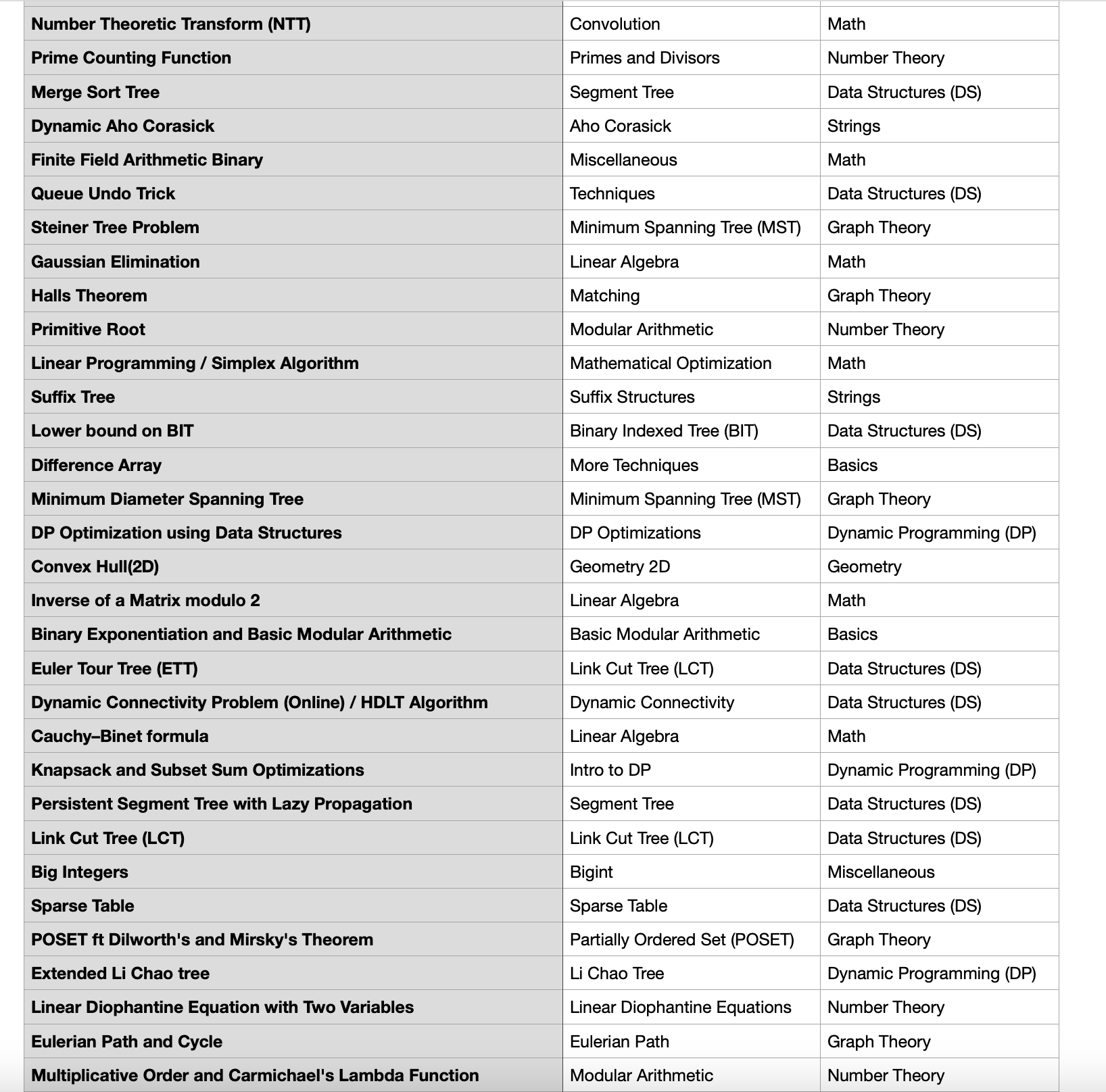}
    \caption{Examples of collected algorithmic tags with their sub-category and category labels.}
    \label{fig:tags}
\end{figure*}

AlgoSimBench contains high-quality human-labeled algorithm tags. We show a portion of them with their subcategory and broad category labels in Figure \ref{fig:tags}.

\subsection{Effects of MCQ Option Position}
\begin{figure}
   \centering
   \includegraphics[width=0.8\linewidth]{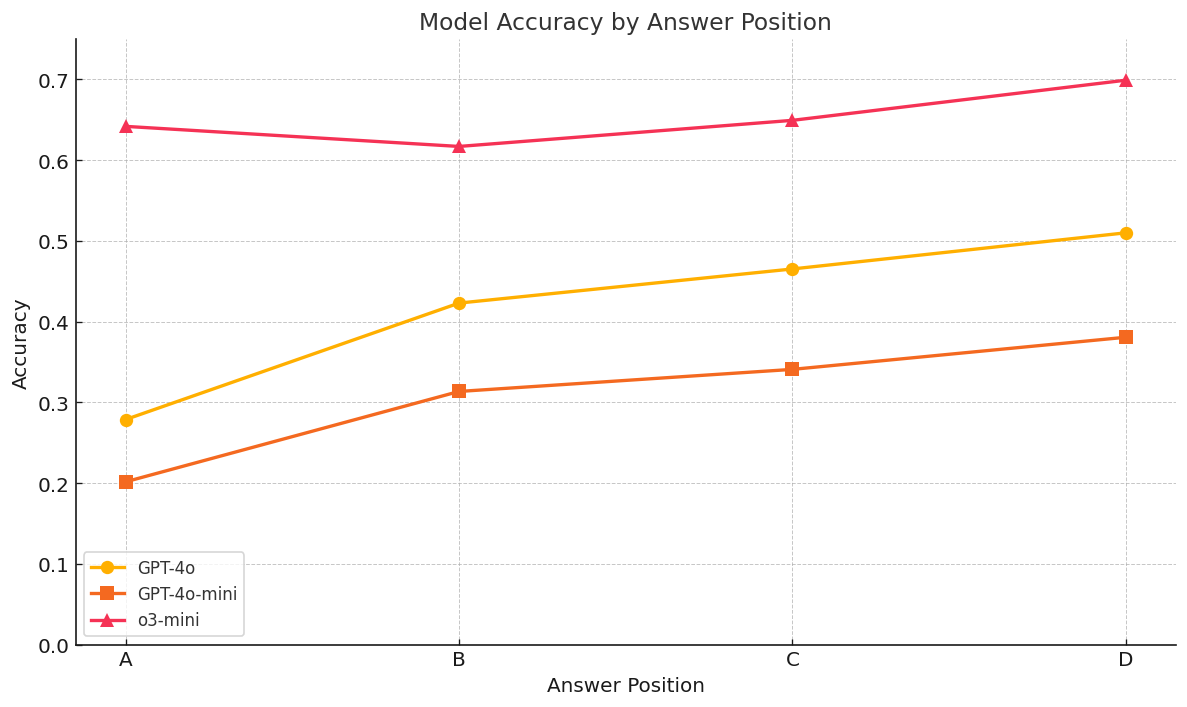}
   \caption{MCQ accuracy if the correct option is fixed to a particular position.}
   \label{fig:option}
\end{figure}

As pointed out in \cite{zheng2024largelanguagemodelsrobust}, large language models are not robust selectors. Therefore, we construct AlgoSimBench MCQs with a random permutation of options, making sure that the correct answer appears in each position with the same probability. To further study whether LLMs might be biased towards selecting particular positions, we experimented with putting the correct answer in a particular position: A, B, C or D. 
The results in Figure \ref{fig:option} show that LLMs tend to select later choices, further stressing the necessity of randomly permuting options.

We keep the random seed = 1000 in our experiments. While the results are generally robust under different random seeds as shown in \ref{tab:seed}
\begin{table*}[t]
\centering
\begin{tabular}{lccccc}
\toprule
\textbf{Model / Input} & \textbf{seed1} & \textbf{seed2} & \textbf{seed3} & \textbf{Seed=1000} & \textbf{Mean $\pm$ Std} \\
\midrule
GPT-4o-mini (statement) & 33.5\% & 34.0\% & 36.8\% & 35.5\% & 35.0\% $\pm$ 1.5\% \\
GPT-4o-mini (ASM-NL)    & 43.3\% & 43.0\% & 43.5\% & 43.8\% & \textbf{43.4\% $\pm$ 0.3\%} \\
GPT-4o (statement)      & 41.0\% & 40.3\% & 42.5\% & 41.5\% & 41.3\% $\pm$ 0.9\% \\
GPT-4o (ASM-NL)         & 53.9\% & 50.2\% & 54.7\% & 53.2\% & \textbf{53.0\% $\pm$ 2.0\%} \\
\bottomrule
\end{tabular}
\caption{Performance under different random seeds for MCQ option layout.}
\label{tab:seed}
\end{table*}

\subsection{Effects of Difficulty level}
\begin{figure}
   \centering
   \includegraphics[width=0.8\linewidth]{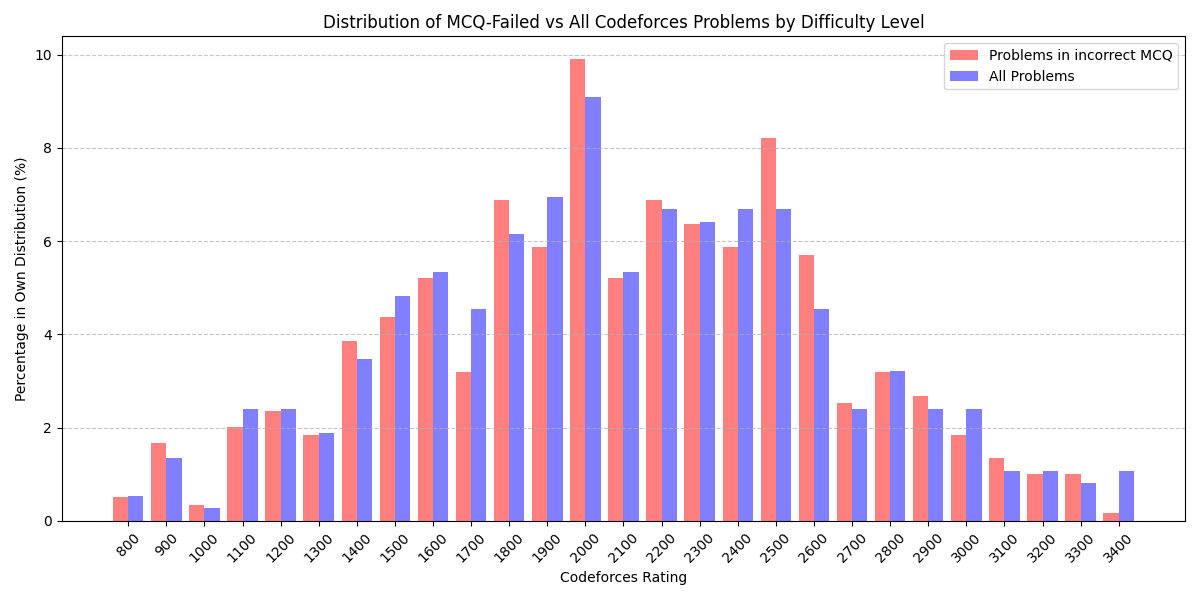}
   \caption{Difficulty-level distribution comparison of failed problems and all problems.}
   \label{fig:difficulty-level distribution}
\end{figure}

As we discussed in the experiments section, the difficulty level rarely correlates with models' performances in End2End-Selection. We provide the distribution comparison of failed problems and all problems in Figure \ref{fig:difficulty-level distribution}.

\subsection{Inference Time}

We also estimate the efficiency of different methods by evaluating the total number of inference tokens needed to obtain the final data. The results are in Table \ref{tab:model_token_inference}, where ``Statement'' and ``Solution'' include the tokens needed to process the corresponding MCQ, while the other three methods include the cost of both processing the MCQ as well the preliminary step of using the LLM to generate the summary or first attempt solution. The cost of ASM-NL is 1.5x to 2.6x compared to the original statement baseline, while ASM-PL is less efficient with an inference time that is 2.6x to 5.3x. Since the accuracy of ASM-NL is about the same or higher than that of ASM-PL, this demonstrates the advantages of generating natural language solutions to help identify ASPs.

\begin{table*}[h]
\centering
\begin{tabular}{lccccc}
\toprule
\textbf{Model} & \textbf{Statement} & \textbf{Summary} & \textbf{ASM-NL} & \textbf{ASM-PL} & \textbf{Solution$^*$} \\
\midrule
GPT-4o-mini & 367 & 788 ({x2.1}) & {937 ({x2.6})} & 1366 ({x3.7}) & 389 \\
GPT-4o & 375 & 667 ({x1.8}) & {857 ({x2.3})} & 1366 ({x3.6}) & 420 \\
o3-mini-medium & 2136 & 2770 ({x1.3}) & 3180 ({x1.5}) & {11268 ({x5.3})} & 2012 \\
Deepseek-R1 & 8381 & 9467 ({x1.1}) & 13371 ({x1.6}) & {28020 ({x3.3})} & 6988 \\
Deepseek-V3 & 511 & 803 ({x1.6}) & {1087 ({x2.1})} & 1845 ({x3.6}) & 483 \\
Claude-3.5-Sonnet & 297 & 537 ({x1.8}) & {588 ({x2.0})} & 1269 ({x4.3}) & 278 \\
Gemini 2.0 Flash & 394 & 654 ({x1.7}) & {997 ({x2.5})} & 1023 ({x2.6}) & 327 \\
\midrule
\textbf{Avg} & 1780 & 2241 ({x1.3}) & 2974 ({x1.7}) & 6594 ({x3.7}) & 1557 \\
\bottomrule
\end{tabular}

\caption{Number of inference tokens on AlgoSimBench-MCQ over different models and methods. \textit{Solution} is a gold-input setting where oracle code solutions were directly provided to the model. ASM-NL and ASM-PL stand for attempted solution matching in natural language and programming language, respectively. The numbers in parentheses show the relative additional cost compared to just processing the original problem statements.}
\label{tab:model_token_inference}
\end{table*}

\subsection{Experimental Settings}
\subsubsection{Prompts}
\begin{figure*}
    \centering
    \includegraphics[width=0.9\linewidth]{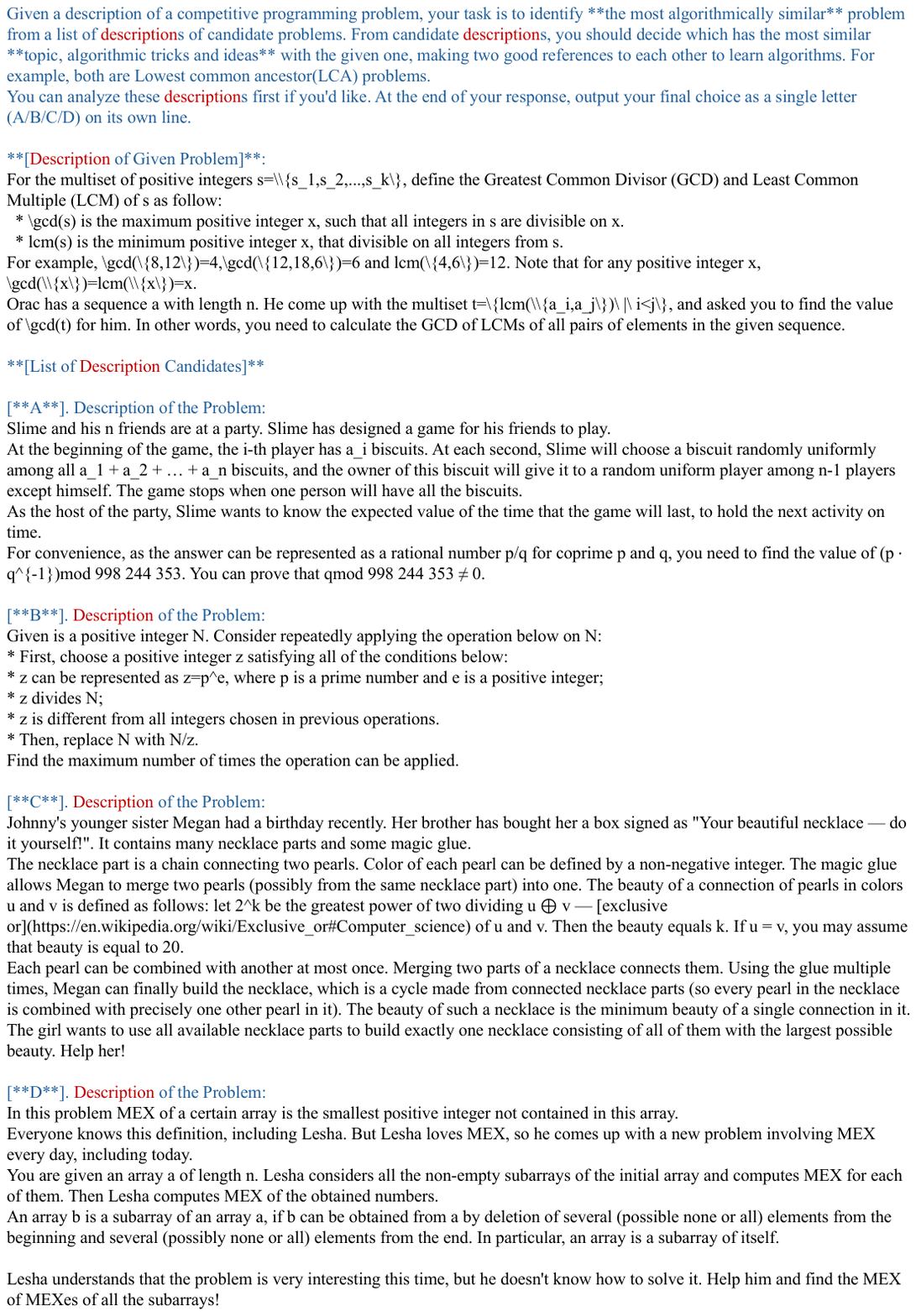}
    \caption{MCQ End2End Prompt with an example from the dataset. (Test cases in problems are omited for presentation.) Blue text is the pre-defined prompt, and red text is highlighted keywords to be replaced for different methods(i.e., ASM/Statement/Summary/Solution).}
    \label{fig:prompt}
\end{figure*}

\paragraph{MCQ prompt} To evaluate models' performances in our experiments (i.e., Table \ref{tab:model_performance}), we apply a simple prompt for all models and methods, as shown in Figure \ref{fig:prompt}. The red-highlighted keywords correspond to our method as below(``keyword''$\rightarrow$method): 
``description''$\rightarrow$ statement; 
``english solution''$\rightarrow$ ASM-NL; ``solution program''$\rightarrow$ solution / ASM-PL.
``problem abstract'' $\rightarrow$ summary.

\paragraph{ASM prompt} ASM methods requires an initial attempt to each problem given by an LLM. To ask the model to give an NL or PL attempted solution, or summarize the problem, we use the simple yet clear instruction:

\textit{Your goal is to give} \{\textit{Problem Restatement/Natural Language Solution/Solution Program\} for a competitive programming problem. Requirements are enclosed in} \{\}. \textit{You may include any thought process as you like, after which, the actual response should strictly follow the formats below:}

\textit{Problem Restatement: }\{\textit{Translate and Abstract the problem into pure mathematical and formal descriptions, focus on the structure of problem rather than input/output/background story details.}\}

\textit{Natural Language Solution: }\{\textit{Analyze and Describe how to approach and solve this problem using English and equations if needed. (E.g., This problem is to find.... it's a classic subset sum problem, ... use 0-1 knapsack... we first...)}\}

\textit{Solution Program: }\{\textit{Implement the Python solution. Wrap your code with ```.}\}

To experiment with ICL, we use the original prompts from the USACO work \cite{shi2024can}.

\subsubsection{Model Details}

We evaluated all LLMs through their official or hosted API, details of the source models used are:
\begin{itemize}
    \item GPT-4o-mini: OpenAI's official API gpt-\textit{4o-mini-2024-07-18}
    \item GPT-4o: OpenAI's official API \textit{gpt-4o-2024-08-06}
    \item o3-mini: OpenAI's official API \textit{o3-mini-2025-01-31}, reasoning=``medium''
    \item Gemini-2.0-Flash: Google AI Studio official API.
    \item Deepseek-V3: OpenRouter's hosted API  \textit{Deepseek-V3-03-24}
    \item Deepseek-R1: OpenRouter's hosted API \textit{Deepseek-R1}
    \item Claude-3.5-sonnet: Anthropic's official API \textit{claude-3-5-sonnet-20241022}
    \item Qwen3-\*-thinking: HuggingFace, \textit{Qwen/Qwen3-0.6B}; \textit{Qwen/Qwen3-4B}; \textit{Qwen/Qwen3-8B}
    \item Llama3.1-8B-Instruct: HuggingFace, \textit{meta-llama/Llama-3.1-8B-Instruct}
    \item Qwen3-Coder-480B: OpenRouter,\textit{ qwen/qwen3-coder}
\end{itemize}

\paragraph{Retrieval Method Details \ref{sec:retri}}

\begin{itemize}
        \item BM25 \cite{trotman2014improvements}, we use BM25Okapi with b=0.75, k1=1.2. 
        \item TF-IDF \cite{10.5555/106765.106782}, we use default parameters without setting thresholds. 
        \item BERT \cite{devlin2019bert}, we use \textit{google-bert/bert-base-uncased} from huggingface, model size 110M parameters. 
        \item BART\cite{lewis2020bart}, we use \textit{facebook/bart-base} from huggingface, model size 139M parameters. 
        \item CodeBert \cite{feng-etal-2020-codebert}, we use \textit{microsoft/codebert-base} from huggingface, model size 125M parameters.
        \item GraphCodeBert \cite{guo2021graphcodebert}, we use \textit{microsoft/graphcodebert-base} from huggingface, model size 125M parameters. 
        \item CodeSage-v2 \cite{zhang2024code}, we use `codesage/codesage-base-v2`, which has 356M parameters.
        \item SFR-Embedding-Code \cite{liu2024codexembedgeneralistembeddingmodel} is from huggingface \textit{Salesforce/SFR-Embedding-Code-400M\_R}, which has 400M parameters.
        \item Jina-Code-V2 \cite{günther2024jinaembeddings28192token} is from huggingface \textit{jinaai/jina-embeddings-v2-base-code} model, which has 161M parameters.
        \item CodeRankEmbed\cite{suresh2025cornstackhighqualitycontrastivedata} is from huggingface \textit{nomic-ai/CodeRankEmbed}, which has 137M parameters.
\end{itemize}

\end{document}